\documentclass[conference]{IEEEtran}
\usepackage{color,latexsym,amsfonts,amssymb}
\usepackage{amsmath,amsthm,cite}
\usepackage[mathscr]{euscript}
\usepackage{graphicx}

\newcommand{\E}{\mathbf{E}}

\newcommand{\Prob}{\mathbf{P}}
\renewcommand{\P}{\mathbf{P}}
\newcommand{\Ind}{\mathbf{1}}
\newcommand{\R}{\mathbb{R}}

\newcommand{\quality}{{q}}
\newcommand{\diversity}{{v}}
\newcommand{\statespace}{\mathcal{S}}
\newcommand{\palmset}{\mathcal{T}}

\newcommand{\event}{\psi}
\newcommand{\lattice}{\statespace}

\newcommand{\region}{\mathcal{R}}

\newtheorem{Theorem}{Theorem}[section]

\newtheorem{Example}[Theorem]{Example}
\newtheorem{Remark}[Theorem]{Remark}

\begin{document}

\title{Determinantal thinning of point processes with network learning applications}

\author{\IEEEauthorblockN{B. B{\l}aszczyszyn and H.P. Keeler}
\IEEEauthorblockA{Inria/ENS, France}}
\date{\today}
\maketitle

\begin{abstract}
A new type of dependent thinning for point processes in continuous space is proposed, which leverages the advantages of determinantal point processes defined on finite spaces and, as such, is particularly amenable to statistical, numerical, and simulation techniques. It  gives a new point process that can serve as a network model exhibiting repulsion. The  properties and functions of the new point process, such as moment measures,  the Laplace functional, the  void probabilities, as well as conditional (Palm) characteristics can be estimated accurately by simulating the underlying (non-thinned) point process, which can be taken, for example, to be Poisson. This is in contrast (and preference to) finite Gibbs point processes, which, instead of thinning, require weighting the Poisson realizations, involving usually intractable normalizing constants. Models based on determinantal point processes are also well suited for statistical (supervised) learning techniques, allowing the models to be fitted to observed network patterns with some particular geometric properties. We illustrate this approach by imitating with determinantal thinning  the well-known Mat{\'e}rn~II hard-core thinning, as well as a soft-core thinning depending on nearest-neighbour triangles. These two examples demonstrate how the proposed approach   can lead to new, statistically optimized, probabilistic transmission scheduling schemes.
\end{abstract}

\section{Introduction}
Researchers have used point processes on the plane to build spatial random models of various wireless network types, but the overwhelming majority  of these models relies upon the Poisson point process~\cite{book2018stochastic}. To develop a more realistic model, while still keeping it tractable, we propose a thinning operation using \emph{discrete determinantal point processes}. 

Originally called fermion point processes by Macchi~\cite{macchi1975coincidence}, determinantal point processes have attracted considerable attention in recent years due to their interesting mathematical properties~\cite{hough2006determinantal}. These point processes admit analytic approaches to several fundamental characteristics such as the Laplace functional, the void probabilities and  Palm distributions~\cite{shirai2003random1}.
They provide useful statistical models for point pattern exhibiting repulsion~\cite{lavancier2015determinantal,biscio2016quantifying} and, compared to the well-studied Gibbs point processes~\cite{dereudre2017introduction}, have advantages such as faster simulation methods and more tractable expressions for likelihoods and moments~\cite{lavancier2015determinantal,lavancier2014detextended}. This has motivated researchers to use these point processes, when defined on the plane $\R^2$, as spatial models for base stations in cellular network~\cite{nakata2014spatial,torrisi2014large,li2014fitting,li2015statistical,gomez2015case}. 

Determinantal point processes are defined usually via factorial moment measures admitting densities in the form of determinants of  matrices populated with the values of some {\em kernel function}. But the main obstacle preventing more use of determinantal point processes in $\R^2$ (or $\R^d$) is the difficulty of finding appropriate kernel functions, which need to define (integral) operators with eigenvalues in the interval~$[0,1]$.  This problem can be largely circumvented when one considers determinantal point processes defined on spaces with finite cardinality, such as bounded lattices, reducing the mathematical technicalities down to  problems of linear algebra. Furthermore, this approach allows the use of non-normalized kernels, which we refer to as {\em $L$-matrices}, to more easily define determinantal processes. In this setting, Kulesza and Taskar~\cite{kulesza2012determinantal} used these point processes to develop a comprehensive framework for statistical (supervised) learning; also see~\cite{kulesza2010structured,kulesza2012arxiv}. 

We leverage this line of research and define point processes in continuous space using a doubly stochastic approach. First, an underlying  point process in a bounded subset $\region\subset\R^d$ is considered, for which a natural  choice  is the Poisson point process. Then, the points of a  given realization of this process are considered as a finite, discrete state space on which a determinantal process (subset of the realization) is sampled using some kernel that  usually depends on the underlying realization. This operation, which  can be seen as a dependent thinning, leads to a new point process existing on bounded regions of $\R^d$ and exhibiting more repulsion than the underlying point process.
Conditioned on a given realization of the underlying point process, the subset point process inherits all closed-form expressions available for discrete determinantal point processes, thus allowing one to accurately estimate the characteristics of the new (thinned) point process by simulating the underlying (non-thinned) point process. The statistical learning approach proposed by Kulesza and Taskar~\cite{kulesza2012determinantal} can then be used to fit the kernel of the determinantal thinning to various types of observed network models.

The paper is structured as follows. In Section~\ref{s.detpp} we recall the basics of the determinantal processes in  finite spaces; in Section~\ref{s.Det-thinning} we introduce the determinantally-thinned point processes and some of their characteristics including  Palm distributions; in Section~\ref{s.fitting} we present the  fitting method based on maximum likelihoods; we demonstrate the results  with two illustrative examples 
 in Section~\ref{s.Cases}; and in Section~\ref{s.Applications} we discuss  network applications.
The code for all numerical results is available online~\cite{keeler2018detpoissoncode}. 

\section{Determinantal point processes}
\label{s.detpp}
We start by detailing determinantal point processes in a discrete setting.
\subsection{Finite state space}
We consider an underlying \emph{state space} $\statespace$ on which we will define a point process (the term \emph{carrier space} is also used). We assume the important simplification that the cardinality of the state space $\statespace$  is finite, that is $\#(\statespace)< \infty$. We consider a simple point process $\Psi$ on the state space $\statespace$, which means that $\Psi$ is a random subset of the state space $\statespace$, that is $\Psi\subseteq \statespace$. A single realization $\psi$ of this point process $\Psi$ can be interpreted simply as occupied or unoccupied locations in the underlying state space $\statespace$. 

\subsection{Definition}
For a state space $\statespace$ with finite cardinality $m:=\#(\statespace)$, a discrete point process is a  determinantal point process $\Psi$ if for all configurations (or subsets) $\event \subseteq  \statespace $, 
\begin{equation}\label{e.dpp}
\Prob(\Psi\supseteq  \event  ) = \det(K_{\psi}),
\end{equation}
where $K$ is some real symmetric $m\times m$ matrix, and $K_{\event}:=[K]_{x,y\in {\event}}$ denotes the restriction of $K$ to the entries indexed by the elements or points in $\event$, that is $x, y\in \psi$.  The matrix $K$ is called the \emph{marginal kernel}, and has to be positive semi-definite. The eigenvalues of $K$ need to be bounded between zero and one  

To simulate or sample a determinantal point process on a finite state space, one typically uses an algorithm based on the eigenvalues  and eigenvectors of the matrix $K$. The number of points is given by Bernoulli trials (or biased coin flips) with the probabilities of success being equal to the eigenvalues, while the  joint  location  of  the  points is determined by the eigenvectors corresponding to the successful trials. Each point is randomly placed one after the other; for further details, see~\cite[Algorithm 1]{kulesza2012determinantal}\cite[Algorithm 1]{lavancier2015determinantal} and \cite[Algorithm 1]{wachinger2015sampling}. 

\subsection{$L$-ensembles}\label{ss.L}
In the finite state space setting kernels~$K$ can be easily defined by using the formalism of $L$-ensembles. Instead of finding a $K$ matrix with appropriate eigenvalues, we can work with a family of point processes known as $L$-ensembles that are defined through a positive semi-definite matrices $L$, which is also indexed by the elements of the space $\statespace$, but the eigenvalues of $L$, though non-negative, do not need to be less than one.
Provided $\det(I+L)\not=0$, where $I$ is a $m\times m$ identity matrix, we define  the kernel
\begin{equation}\label{e.K-L}
K=L(I+L)^{-1}.
\end{equation}
This mapping~\eqref{e.K-L} preserves the eigenvectors and maps the corresponding eigenvalues by the function $x/(1+x)$. Consequently, the  kernel $K$ given by~\eqref{e.K-L} is positive semi-definite with eigenvalues between zero and one. The corresponding determinantal point process $\Psi$   
satisfies
\begin{equation}\label{e.dpp-L}
\Prob(\Psi= \event  ) = \frac{\det(L_\psi)}{\det(I+L)}.
\end{equation}
The relation~\eqref{e.K-L} can be inverted yielding the $L$-ensemble representation of the determinantal point process
\begin{equation}\label{e.L-K}
L=K(I-L)^{-1},
\end{equation}
provided all eigenvalues of $K$ are strictly positive, which is equivalent to $\P(\Psi=\emptyset)>0$. For more details, see, for example, the paper by Borodin and Rains~\cite[Proposition 1.1]{borodin2005eynard} or the book~\cite{kulesza2012determinantal} by Kulesza and Taskar.

\section{Determinantally-thinned point processes}
\label{s.Det-thinning}
We now define a new point process, which  builds upon a homogeneous Poisson point process $\Phi$ with intensity $\lambda>0$ on a bounded region $\region\subset \R^d $ of the $n$-dimensional Euclidean space. Given a realization $\Phi= \phi$, we  consider it as the state space $\statespace=\phi$ on which a determinantal point process (subset)  $\Psi\subset\phi$ is sampled,  resulting in a (typically dependent) thinning of the realization $\Phi= \phi$.  More precisely,  the points in a realization $\phi=\{x_i\}_i$ of a Poisson point process $\Phi$ form the state space of the finite determinantal point processes $\Psi$, which is defined via 
\begin{equation}\label{e.dppp}
\Prob(\Psi\supseteq  \event |\Phi=\phi ) = \det(K_{\psi}(\phi)),
\end{equation}
where $K_{\event}(\phi)=[K(\phi)]_{x_i,x_j\in {\event}}$ is such that $\psi\subset \phi\subset {\cal{R}} $. 
Note that $\Psi$ is characterized by intensity measure of the underlying Poisson point process $\Phi$ and  the function $K(\cdot)$
which maps each (Poisson)  realization $\phi$ to  a semi-definite matrix $K(\phi)$ (with eigenvalues in $[0,1]$) having elements indexed by the points of $\phi$.

The point process  {$\Psi $} is defined on a subset of~{$\R^d$}, but uses the discrete approach of determinantal point processes. In other words, the points of the realization {$\phi$} are dependently thinned such that there is repulsion among the points of  {$\Psi$}. We call this point process a \emph{determinantally-thinned Poisson point process} or, for brevity, a \emph{determintantal Poisson process}.  

The double stochastic construction of determinantally-thinned point processes can be compared with the classic \emph{Mat\'ern hard-core processes} (of type I, II and III), which are also constructed through dependent thinning of underlying Poisson point processes.  For these point processes, there is a zero probability that any two points are within a certain fixed distance of each other. Determinantal thinning of Poisson point processes can provide examples of soft-core process, where there is a smaller (compared to the Poisson case) probability that any two points are within a certain distance of each other. We return to this theme later in our results section, where we fit our new point process to a Mat\'ern~II hard-core process. 

\subsection{Functionals of $\Psi$}
The double stochastic construction of $\Psi$ gives
\begin{equation}
\E[h(\Psi)]= \E [\E[  h( \Psi) |\Phi] ],
\end{equation}
where $h$ is a general real function on the space of realizations of point processes on $\mathcal{R}$ (measurable with respect to the usual $\sigma$-algebra of counting measures), and  the conditional expectation on the right-hand-side can be calculated using~\eqref{e.dppp}. Several special cases of $h$ admit explicit expressions for this conditional expectation allowing one to express $\E[h(\Psi)]$ in terms of some other functional of the Poisson point process $\Phi$. 
The evaluation of such expressions requires simulating at most the underlying Poisson point process~$\Phi$, but not $\Psi$. This not only simplifies the task but also reduces the variance.

\subsubsection{Average retention probability}
We consider the average probability that a point is retained (or not removed) after the thinning. The average retention probability of  a Poisson point located at $x\in {\cal{R}}$ is~\footnote{To simplify the expressions we slightly abuse the  notation writing $\phi\cup x$ instead of $\phi\cup\{x\}$.} 
\begin{align}\label{e.pi}
\pi(x):= \E[\P( x \in \Psi |\Phi\cup x )] 
=
\E\left[K_{x}(\Phi\cup x) \right]
\end{align}
where $K(\Phi\cup x)$ denotes the kernel matrix with the entries corresponding to the state space $\statespace={\Phi\cup x}$, and $I$ is an identity matrix with cardinality of $(\Phi\cup x)$, that is $\Phi({\cal{R}})+1$, and $K_{x}(\Phi\cup x)=K_{\{x\}}(\Phi\cup x)$ denotes the 
restriction  of $K(\Phi\cup x)$ to a single element on the diagonal 
corresponding to $x$.

\subsubsection{Moment measures and correlation functions}
Multiplying the intensity measure~$\Lambda(d x)$ of the underlying Poisson process by the average retention probability, we obtain the intensity measure of $\Psi$, namely $M(d x):=\pi(x)\Lambda(dx)$. 
In the case of underlying homogeneous Poisson process of intensity $\lambda$ (within the considered finite window)
we  can express the first moment measure as
\begin{align}\label{e.mean-measure}
M(B):=\lambda \int_B \pi(x)  dx&=\lambda|B|\E  \left[\det[K_{U}(\Phi\cup U) \right], 
\end{align}
where $B\subset \region$, $|B|$ is the area of $B$, and  $U$ is a single point uniformly located in $B$. 

The  higher \emph{factorial moment measures} are similarly given by
\begin{equation}\label{e.moments}
M^{(n)}(d(x_1,\dots,x_n))= \pi(x_1,\dots, x_n)\Lambda(d x_1)\ldots\Lambda(d x_n)
\end{equation}
where
\begin{equation}\label{e.pin}
\pi(x_1,\ldots,x_n)=\E\left[\det[K_{(x_1,\dots, x_n)}(\Phi\cup (x_1,\dots, x_n))] \right]\,
\end{equation}
for $x_1\neq,\dots,\neq x_n$ and $0$ otherwise, 
and $K_{(x_1,\dots, x_n)}(\Phi\cup (x_1,\dots, x_n))$ is the restriction of the matrix $K(\Phi\cup (x_n,\dots, x_n))$ to the 
the elements $x_1,\dots, x_n$. 
 
For the the case of an underlying homogeneous Poisson process with intensity $\lambda$, the  second factorial moment measure  can be written as
\begin{align}
M^{(2)}(B_1,B_2)
&=\lambda^2|B_1||B_2|\E  \left[K_{(U,V)}(\Phi\cup U\cup V)]  \right],
\end{align}
where $U$ and $V$ are points uniformly located in $B_1$ and $B_2$ respectively, and a similar expression holds for  $n$-th moment measure. These expressions allow us to more efficiently estimate the measure with stochastic simulation.

\begin{Remark}
When the matrix $K_{(x_1,\dots, x_n)}(\Phi\cup (x_n,\dots, x_n))$ in~\eqref{e.pin} 
does not depend on $\Phi$ and   $[K_{(x_1,\dots, x_n)}]_{x_i,x_j}=\mathcal{K}(x_i,y_j)$ for some appropriate function $\mathcal{K}$, then 
$\Psi$ has the moment measures~\eqref{e.moments} in the form of a (usual, continuous) determinantal point process with kernel~$\mathcal{K}$. However, it does not seem evident how one finds $\mathcal{K}$ such that the resulting  matrix $K$ has eigenvalues in~$[0,1]$ for all configurations of points $(x_1,\dots,x_n)$, $n\ge 1$. Note that this is not equivalent to guaranteeing that the  eigenvalues of the integral operator related to $\mathcal{K}$ are in~$[0,1]$,
the latter condition being usually required in continuous determinantal point process framework. 
\end{Remark}

\subsubsection{Void probabilities and the complement process}
We recall that the number of points of a determinantal point process is equal (in distribution) to the number of successful
   Bernoulli trials with the eigenvalues of~$K$ as the parameters. Using this result, one can show that the probability of no point of $\phi$ 
being retained for $\Psi$ is given by $\P(\Psi\cap\psi=\emptyset)=\prod_i(1-\lambda_i(\phi))=\det((I-K)_\phi)$, where $\lambda_i(\phi)$ are the eigenvalues of $K_\phi$.
Consequently, we can express the void probabilities of the determinantal thinning $\Psi$ of $\Phi$ by the following expression
\begin{align}
\nu_\Psi(B)&:=\P(\Psi(B)=0)\nonumber\\
&=\E[\det((I-K(\Phi))_{\Phi\cap B}].\label{e.void}
\end{align}
Observing again  in the discrete setting of $\statespace$ that $\P(\statespace\setminus\Psi\supset \psi)=\P(\Psi\cap\psi=\emptyset)=\det((I-K)_\phi)$
one sees easily that the point process $\Psi^c:=\Phi\setminus\Psi$ formed from the Poisson points removed by the determinantal thinning with kernel $K(\cdot)$ is also a determinantally-thinned Poisson process with the kernel  $I-K(\cdot)$. (But the retained and removed points are not in general independent of each other as in the case of an independent thinning of Poisson processes.)

\subsubsection{Laplace functional}
For any non-negative function $f$, the Laplace functional of the detetermintally-thinned Poisson process $\Psi$ is given by
\begin{align}\nonumber
\textbf{L}_{\Psi}(f):&=\E\,\left[e^{-\sum_{x\in \Psi} f(x) } \right] \\
&=\E\,\left[\det[I-K'(\Phi) \right] ,\label{e.Laplace}
\end{align}
where the matrix $K'=K'(\phi)$ has the elements 
\begin{equation}\label{e.Kernel-Laplace}
[K]_{x_i,y_j}'=[1-e^{-f(x_i)}]^{1/2}\, [K]_{x_i,x_j}\, [1-e^{-f(x_j)}]^{1/2},
\end{equation}
for all $x_i,x_j\in\phi$.
 Shirai and Takahashi~\cite{shirai2003random2} proved this in the general discrete case. But  in the  Appendix~\ref{s.Appendix-Laplace} we present a simpler, probabilistic proof of the last equality, which leverages the finite state space assumption of the determinantal process, circumventing the functional-analytic techniques used by Shirai and Takahashi.

\subsubsection{Palm distributions}
\label{sss.Palm}
Palm distribution  of a point process can be interpreted as the probability distribution of a point process $\Psi$ conditioned on a point of the point process $\Psi$ existing at some location $u$ on the underlying state space $\statespace$. If we condition on $n$ points of the point process existing at $n$ locations $\palmset=\{x_1,\dots, x_n\}$, then we need the $n$-fold Palm distribution. The reduced Palm distribution is the Palm distribution when we ignore (remove) the points in the set~$\palmset$. The reduced Palm version  $\Psi^\palmset$ of $\Psi$  given points at $\palmset$, is a  determinantal thinning of some Gibbs point process having density with  respect to the original  Poisson process $\Phi$. More precisely,
for any  real measurable function $h$ on the space of realizations of point processes on~$\mathcal{R}$,
\begin{equation}\label{e.Palm-T}
\E[h(\Psi^\palmset)]= \frac{1}{\pi(\palmset)}\E[\E[h(\bar\Psi^{\palmset})|\Phi]\det( K_\palmset(\Phi\cup\palmset))],
\end{equation}
where $\pi(\palmset)=\pi(x_1,\dots, x_n)$ is given by~\eqref{e.moments} and 
$\bar\Psi^{\palmset}$
is a determinantal thinning of $\Phi$ 
with Palm kernel $K^\palmset(\phi)$ given by the  Schur complement  of the block (or submatrix)
$K_\palmset(\phi\cup\palmset)$ of the matrix $K(\phi\cup\palmset)$; see Appendix~\ref{s.Appendix-Palm}.
In the case of one-point conditioning
$\palmset=\{u\}$, the corresponding  Schur complement representation of the Palm kernel $K^u=K^u(\phi)$ has elements  
\begin{align}
[K^u]_{x,y}
&=[K]_{x,y}-\frac{[K]_{x,u}[K]_{y,u}}{[K]_{u,u}}\,,\label{e.Palm}
\end{align}
where $K=K(\phi\cup u)$, so $K$ and $K_u$ are $(n+1)\times(n+1)$ and $n\times n$ matrices.

The general expression in the right-hand-side  of~\eqref{e.Palm-T} can be understood in the following way: $\bar\Psi^{\palmset}$ is the reduced Palm version of $\Psi$, on  each realization $\Phi\cup\palmset$ (that is, conditioned to  contain $\palmset$, and  with $\palmset$  removed from
the conditional realization).  
The biasing by   $\det (K_\palmset(\Phi\cup\palmset))$  
transforms conditioning on a  given realization of Poisson process  to the average one. 
We  shall prove~\eqref{e.Palm-T}  
in Appendix~\ref{s.Appendix-Palm}, where  we also recall two further, equivalent characterizations of the reduced Palm distribution~$\bar\Psi^{\palmset}$
in the discrete setting.

\subsubsection{Nearest neighbour distance}
Using~\eqref{e.Palm} and~\eqref{e.void} 
one can express the distribution function $G^u(r)$
of the distance from the point of $u\in\Psi^u$ to its nearest neighbour 
\begin{align}\nonumber
G^u(r)&=1-\P\left(\min_{x\in\Psi^u}|u-x|>r\right)\\
&=1-\nu_{\Psi^u}(B_u(r))\nonumber\\[0.5ex]
&=\frac{\E[(1-\det[(I-K^u(\Phi))]_{\Phi\cap B_u(r)})[K]_{u,u}(\Phi\cup u)]}{\pi(u)},\label{e.NN}
\end{align}
where $\pi(u)$ is the average retention probability~\eqref{e.pi}.

\section{Statistical fitting}\label{s.fitting}
Discrete determinantal point processes are suitable for fitting techniques such as max-log-likelihood-methods~\cite{lavancier2015determinantal,kulesza2012determinantal}. Relevant to the present work, Kulesza and Taskar~\cite{kulesza2012determinantal} developed a statistical (supervised) learning method allowing one to approximate an empirically-observed thinning mechanism by a determinantal thinning model. In other words, the training data consists of sets with coupled subsets that need to be fitted. This approach was originally motivated by the automated analysis of documents (the sets) and generation of their abstracts (the subsets). Inspired by this work, our proposal is to fit determinantally-thinned point processes to real network layouts, with particular focus on models of (transmission) scheduling schemes:  locations of potential transmitters are the underlying point patterns (the sets) and the locations   actually  scheduled for  transmissions are the retained points (the subsets). 

\subsection{Specifying quality and repulsion of points}
For  an interpretation of the $L$-matrix, we briefly recall the approach proposed by Kulesza and Taskar~\cite{kulesza2012determinantal}. Consider a matrix $L$ whose elements can be written as
\begin{equation}\label{e.Ldecomp}
[L]_{x,y} = \quality_x \,[S]_{x,y}\, \quality_y ,
\end{equation}
for $x,y\in\lattice$, where $q_x\in \R^+$  and $S$ is a symmetric, positive semi-definite $m\times m$ matrix. These two terms are known as \emph{quality} and the \emph{similarity matrix}. The quality $q_x$ measures the goodness of point $q_x\in\statespace$, while $[S]_{x,y}$ gives a measure of similarity between points $x$ and $y$.  The larger the $q_x$ value, the more likely there will be a point of the determinantal point process at location~$x$, while the larger  $[S]_{x,y}$ value for two locations $x$ and $y$, the less likely realizations will occur with two points simultaneously at both locations. If $q_x\le 1$, then an additional probabilistic interpretation exists: Determinantal point process $\Psi$ characterized by $L$-ensemble $L$ ($L$-ensemble process for short) in the form of~\eqref{e.Ldecomp} is an independent thinning of the $S$-ensemble, with retention probabilities equal to $q^2_x$, for all $x\in\statespace$.

One way of constructing a positive semi-definite matrix $L$ is to use some known kernel functions, such as those used for covariance functions of Gaussian processes.
\begin{Example}[Squared exponential (or Gaussian) similarity kernel]\label{eg.gausskernel}
For $\statespace\subset\R^d$, the similarity kernel is $[S]_{x,y} =Ce^{-(|x_i-x_j|^2/\sigma^2)}$, where $|\cdot|$ is the Euclidean distance and $\sigma>0 $, $C>0$ are  suitably chosen the parameters. 
\end{Example}
Another possibility is to specify  $S$  as some  Gram matrix 
 $S=B^\top B$, where, often normalized, columns of the matrix $B$ are 
  some vectors   $\diversity_x$
 representing points $x\in\statespace$ in the state space $\statespace$.  Quantities such as these vectors are called \emph{covariates} (in statistics) or \emph{features} (in computer science), among other terms.   
 The dimension of these {\em diversity (covariate or feature)  vectors}  can be arbitrary. Note in this case
the similarity between locations $x$ and $y$ is given by the scalar product of the respective diversity vectors
$[S]_{x,y} = \diversity_x^{\top}  \diversity_y$, thus points with more collinear  diversity vectors repel each other more.

Similarly, the scalar-valued qualities $q_x$ can be modeled by using some {\em quality (covariate or feature) vectors} $f_x$ of some arbitrary dimension.
The following construction  will be used in our  numerical examples.
\begin{Example}[Exponential quality model]
\label{exe.quality}
The qualities $q_x$ depend on the quality vectors $f_x$ in the following way 
\begin{equation}\label{e.quality}
\quality_x=\quality_x(\theta) := e^{\left( {\theta^\top} {f_x}\right) }\,,
\end{equation}
where $\theta$ is a parameter vector with  the same dimension as $f_x$.  
\end{Example}

\subsection{Learning determinantal thinning parameters}
Consider a situation when some number of pairs of patterns of points  $(\phi_t,\psi_t)$, $t=1\ldots,T$ is observed, where $\phi_t$ is a realization of a (say Poisson) process and $\psi_t\subset\phi_t$ is some  subset (due to thinning) of this realization.
Our goal is to fit a determinantal thinning model $\Psi$ to this observed data. More precisely, we will find  $\Psi$ which maximizes the likelihood of observing thinned realizations $\psi_t$, given (complete) realizations $\phi_t$,
assuming independence of realizations of pairs across $t=1,\ldots,T$.
This is equivalent to the maximization of the following log-likelihood
\begin{align}\nonumber
\mathcal{L}_{\{(\phi_t,\psi_t)\}}&=
\log\Bigl(\prod_{t=1}^T\P(\Psi=\psi_t|\Phi=\phi_t)\Bigr)\\
&=\sum_{t=1}^T\log\Bigl(\frac{\det(L_{\psi_t}(\phi_t)}{\det(I+L(\phi_t)}\Bigr)\,,
\label{e.log-like-general}
\end{align}
where $L(\phi)$ is the $L$-ensemble characterizing the determinantal thinning of~$\Phi$. Fitting the determinantal thinning to $(\phi_t,\psi_t)$, $t=1\ldots,T$ consists thus in finding model parameters that maximize~\eqref{e.log-like-general}. The exponential quality model~\ref{exe.quality} with an arbitrary similarity matrix $S$ allows for standard optimization methods because the expression in~\eqref{e.log-like-general} is a concave function of  $\theta$ as shown by  Kulesza and Taskar~\cite[Proposition 4.1]{kulesza2012determinantal}).    

\section{Test cases}\label{s.Cases}
We will illustrate the fitting method of the determinantally-thinned  Poisson process $\Psi$, outlined in  Section~\ref{s.fitting}, by fitting the  point process $\Psi$ to two types of points processes both constructed through dependent thinning. These two point processes are suitable and demonstrative models as they have a similar two-step construction:  1) Simulate a Poisson point process. 2) Given a realization of this point process, retain/thin the points according to some rule. 
We will see that they also  represent,  in some sense, two extreme  cases: one is well captured just by the diversity matrix, the other just by the quality model.

\subsection{Training sets: Two dependently-thinned point process}
\subsubsection{Mat{\'e}rn II case} 
The first test case is the well-known {\em Mat{\'e}rn II} point process. To construct it, all points of the underlying Poisson process are assigned an independent and identically distributed mark, say a uniform random variable on $[0,1]$,
and then the points that have a neighbor within distance $r_{\text{M}}$ with a smaller mark are removed; for details, see, for example, the books~\cite[Section~5.4]{chiu2013stochastic} or \cite[Section~3.5.2]{book2018stochastic}. The Mat{\'e}rn  II model is characterized  by two parameters: an inhibition radius $r_{\text{M}}>0$ and density $\lambda>0$ of the underlying Poisson point process. The resulting density is $ \lambda_{\text{M}} =(1- e^{-\lambda \pi r^2})/( \pi r_{\text{M}}^2)$. 

\subsubsection{Triangle case} 
For the second test case, we remove a given Poisson point if the total distance to its first and second nearest neighbour plus the 
distance between these two neighbours exceeds some parameter $r_{\text{T}}>0$.We refer to the resulting random object simply as a \emph{triangle (thinned) point process}. No explicit expressions will be used for  this process.

\subsection{Quality and diversity models}
In our model of $\Psi$, we assume the quality feature or covariate $f_x(\phi) \in \R^4$ of a point $x$ within a given configuration~$\phi$ to be a four-dimensional  vector composed of a constant, the distances $d_{x,1}$, $d_{x,2}$ of $x$ to its two nearest neighbours, as well as the distance $d_{x,3}$ between these two neighbours. Consequently, the scalar product  $\theta^\top f_x$ in~\eqref{e.quality} is equal to 
\begin{equation}
{\theta^\top} {f_x(\phi)} =\theta_0+\theta_1 d_{x,1} +\theta_2 d_{x,1}+\theta_3 d_{x,3} \, .    
\end{equation}
For our similarity matrix $S$, we use the squared exponential (or Gaussian) kernel given in Example~\ref{eg.gausskernel} with the constant $C=1$,  which means it is also possible to fit the $\sigma$ parameter, thus adjusting the repulsion between points. 

\subsection{Simulation and code details}
To remove edge effects, the two test point processes are built on Poisson point processes simulated in windows that are extended versions of the observation windows used in the fitting stage. For example, if the observation window of the thinned-point processes is a disk with radius $r_{\text{W}}$, then we simulate the underlying Poisson point process on the  disks with radius $r'_{\text{W}}=r_{\text{W}}+r_{\text{M}}$ in the Mat\'ern case  and $r'_{\text{W}}=r+2r_{\text{T}}$
in the triangle case, and then thin the points accordingly. But the fitting (or learning) data will only contain points on the original disk of radius $r_{\text{W}}$, which means that the fitted determinantal thinning will be dependent on the boundary of the observation window. 

We implemented everything in MATLAB~\cite{keeler2018detpoissoncode} and ran it on a standard machine, taking mostly seconds to complete each of the three components:
generation of the test cases; fitting of a determinatally-thinned model; and empirical validation of fitted model.  The fitting method (outlined in Section\ref{s.fitting}) used a standard optimization function (\texttt{fminunc}) in MATLAB that uses the Broyden-Fletcher-Goldfarb-Shanno (BFGS) algorithm, which is a popular type of gradient method, particularly in machine learning. 

\subsection{Numerical results}
We simulated a Poisson point process with intensity  $\lambda=10$, where the circular observation window had a radius $r_{\text{W}}=1$.  For the Mat{\'e}rn II, the parameters were $r_{\text{M}}=0.2530$ (yielding  $\lambda_{\text{M}}=4.3076$). For the triangular process, they were $r_{\text{T}}= 0.6325$ ($\hat{\lambda}_{\text{T}}=4.8961$ an empirical estimate).

\subsubsection{Quality versus diversity fitting}
We found that we could fit the point process $\Psi$ to the Mat{\'e}rn II process by just optimizing $\sigma$ and $\theta_0$, and so ignoring the non-constant terms 
of the quality feature vectors $f_x$. The addition of the terms $\theta_3$ and $\theta_4$ gave negligible gain. This suggests that the essence of the point process Mat{\'e}rn II is captured through its repulsion, and not by our choice of the quality model $q_x(\theta)$. 

Conversely, for the triangle process, we could set $\sigma=0$, thus reducing $S$ to an identity matrix, and still accurately fit our point process $\Psi$ very well by fitting the parameters $\theta_1$ to $\theta_4$. This observation suggests the nature of the triangle model is captured well by its nearest neighbour distances, which of course agrees completely with its construction. In fact, we were able to fit our point process $\Psi$ to the triangle model so well that, at times, the randomness (or variance) of the fitted $\Psi$ decreased significantly, due to the fact that the quality $q_i(\theta)$ features dictated overwhelmingly where points of $\Psi$ should and should not exist. In short, the Mat{\'e}rn II and triangle processes represent well, in some sense, the two model extremes: models captured by just diversity or just quality of points.

Based on a $100$ samples (or training sets) of our two test cases, we were able to find the $\theta$ value that maximized the log-likelihood \eqref{e.log-like-general}, denoted by $\theta^*$. We arrived at the fitting parameters: Mat{\'e}rn II (with fitted $\sigma=1.5679$) is  $\theta^*=(0.3067,0.6315)$ and triangular process (with $\sigma=0$) is $\theta^*=(-4.0779  ,  2.7934  ,  1.2445 ,   2.1173)$. We give examples of realizations the point process $\Psi$ fitted to these two models in Figures~\ref{ConfigMaternII} and~\ref{ConfigTri}. 

\subsubsection{Testing the quality of fit}
\begin{figure}[t]
\begin{minipage}[b]{0.48\linewidth}
\centering
\centerline{\includegraphics[width=1.1\linewidth]{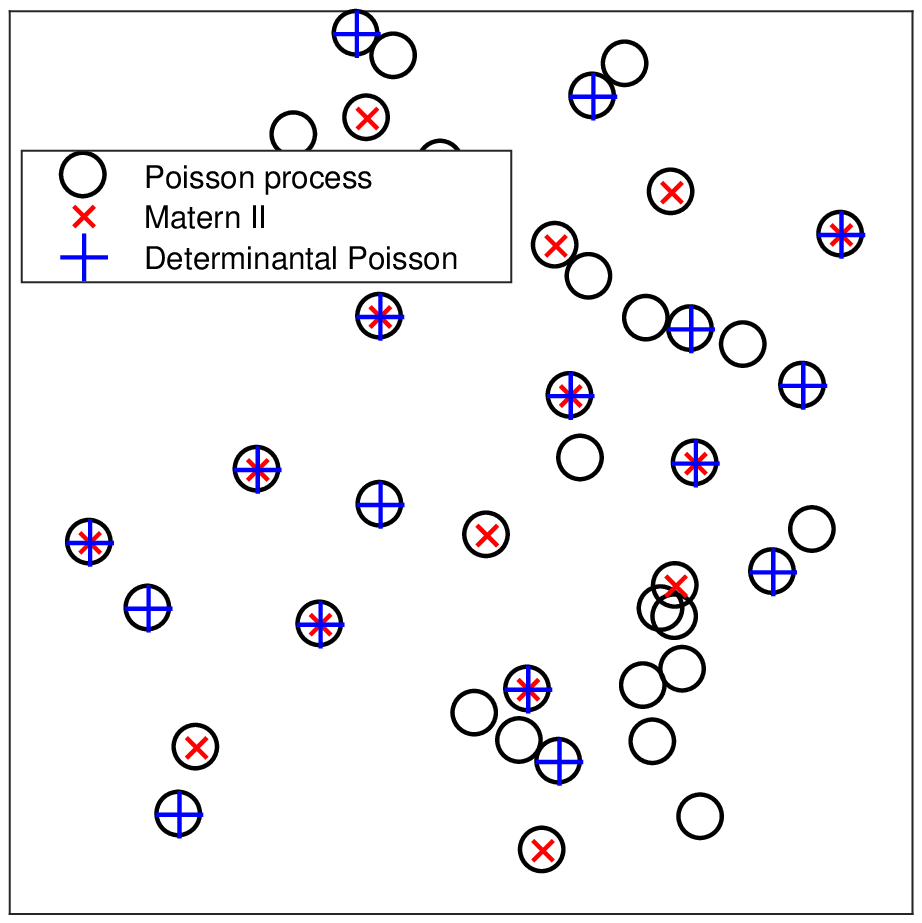}}
\vspace{-2ex}
\caption{\footnotesize Realizations of a  Mat{\'e}rn II process and a fitted determinantally-thinned Poisson  process on a unit disk (both generated  on the same Poisson point process). }
\label{ConfigMaternII}
\end{minipage}
\hspace{0.1em}
\begin{minipage}[b]{0.48\linewidth}
\centering
\centerline{\includegraphics[width=1.1\linewidth]{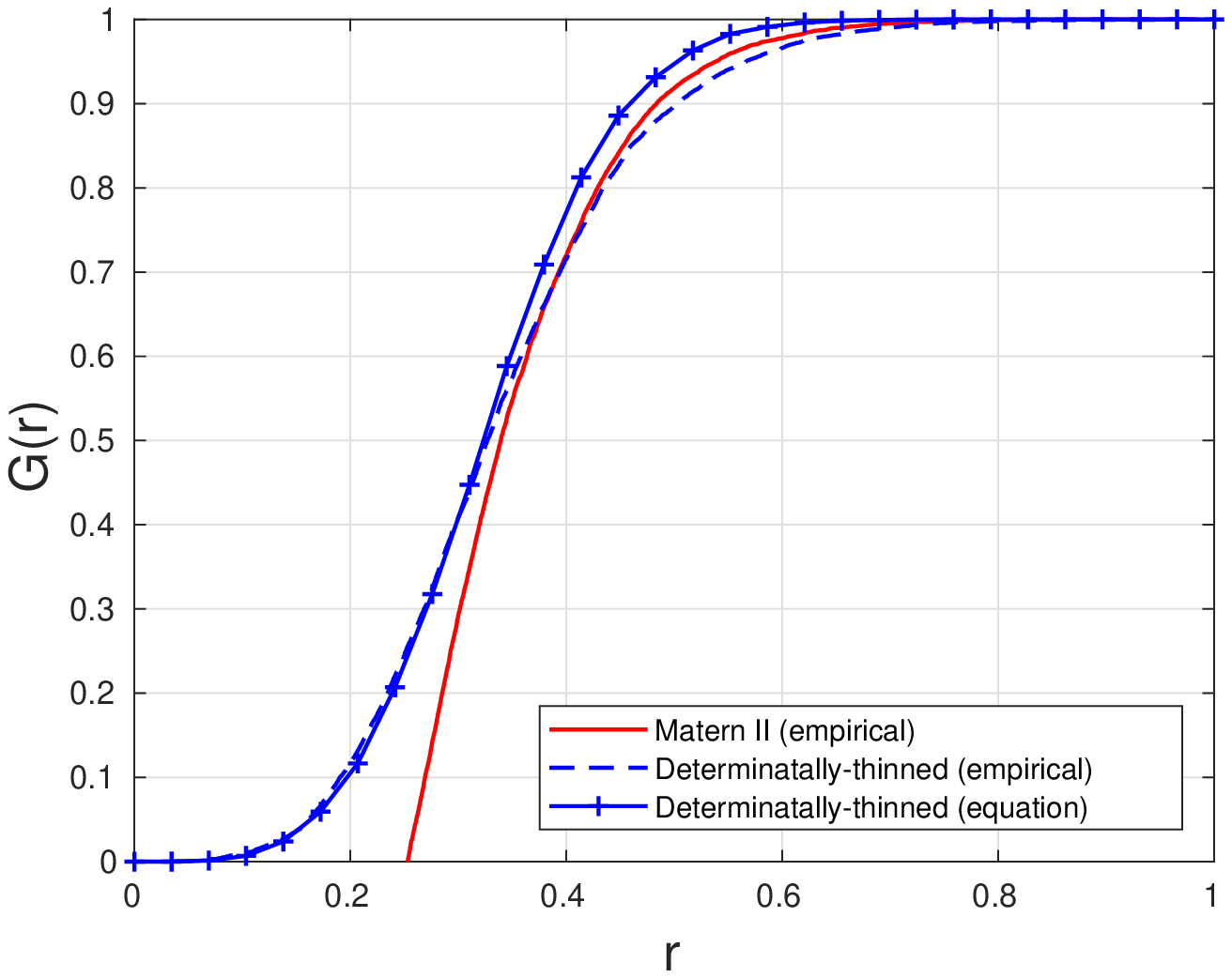}}
\vspace{-2ex}
\caption{\footnotesize
 Nearest-neighbour distributions of the  determinantally-thinned Poisson Poisson process fitted to the Mat{\'e}rn  II process:
empirical (average) and the semi-analytic~\eqref{e.NN} calculated for the point located at the origin $u=o$.}
\label{NeighDistMaternII}
\end{minipage}
\vspace{-3ex}
\end{figure}

\begin{figure}[t]
\begin{minipage}[b]{0.48\linewidth}
\centering
\centerline{\includegraphics[width=1.1\linewidth]{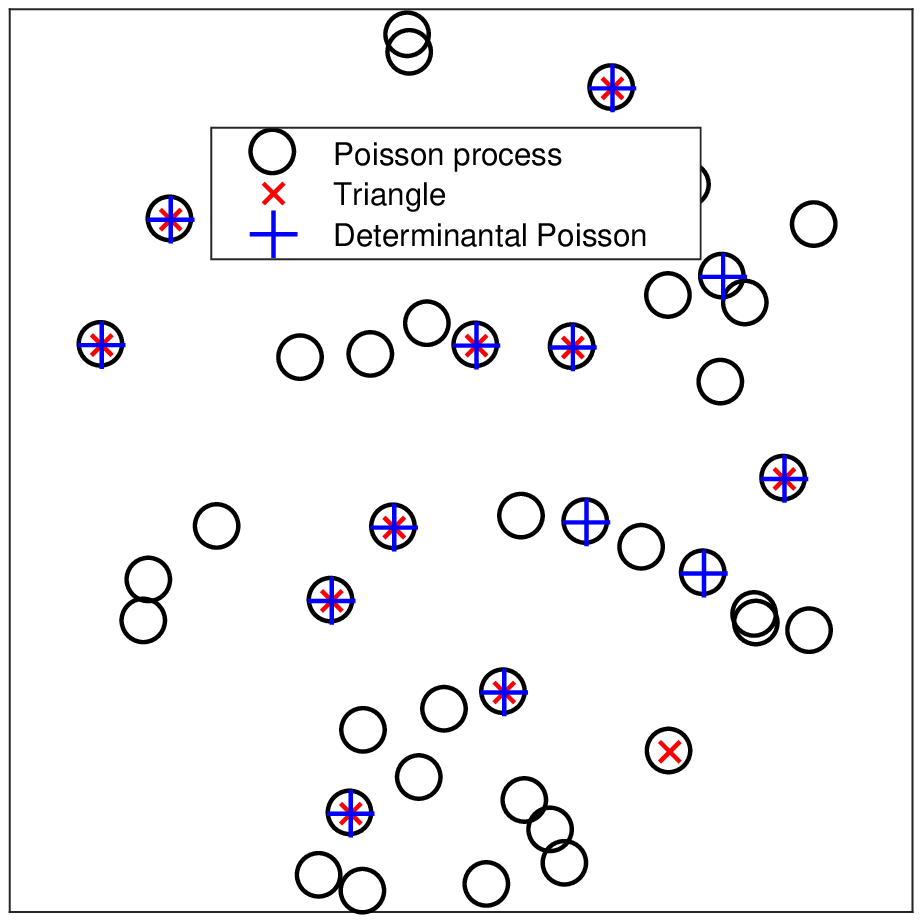}}
\vspace{-2ex}
\caption{\footnotesize Realizations of a triangle thinning process and a fitted determinantally-thinned Poisson  process on a unit disk (both generated  on the same Poisson point process). }
\label{ConfigTri}
\end{minipage}
\hspace{0.1em}
\begin{minipage}[b]{0.48\linewidth}
\centering
\centerline{\includegraphics[width=1.1\linewidth]{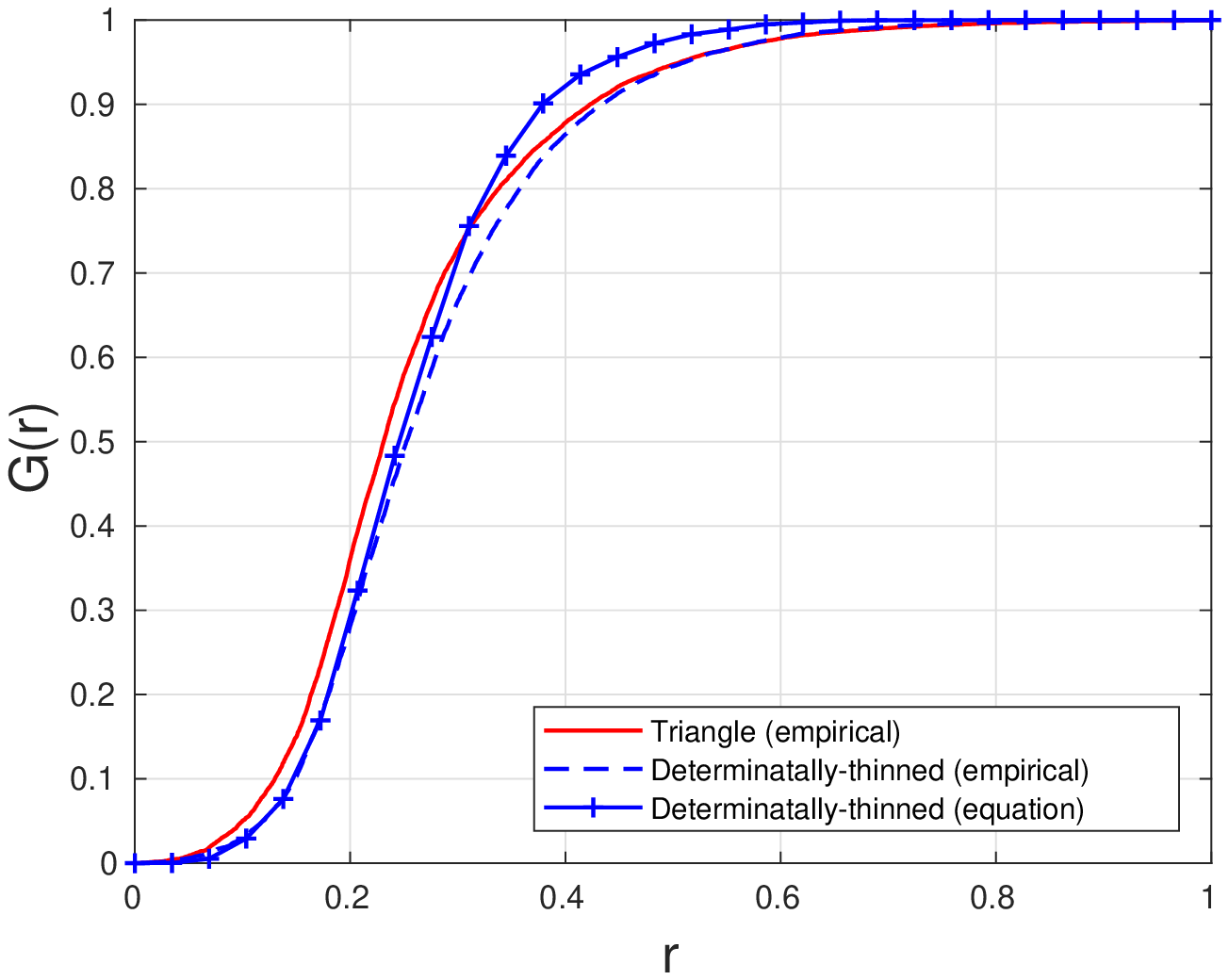}}
\vspace{-2ex}
\caption{\footnotesize  Nearest-neighbour distributions of the  determinantally-thinned Poisson Poisson process fitted to the
triangle thinning process: empirical and semi-analytic as on Figure~\ref{NeighDistMaternII}.}
\label{NeighDistTri}
\end{minipage}
\vspace{-3ex}
\end{figure}

\paragraph{Nearest neighbour distance}
To gauge the quality of the fitted models, we empirically estimated the average nearest neighbour distribution $G(r)$; see . Unfortunately, this quantity is highly susceptible to edge effects, as points near the edge of the observation window generally have less neighbours, which means the empirical estimates of our test cases and (fitted) determinantal models are biased; see Figures~\ref{NeighDistMaternII} and~\ref{NeighDistTri}. But our semi-analytic formula~\eqref{e.NN} does not suffer from this bias, giving an accurate description of the nearest neighbour distribution~$G^u$  for a  point $u$ at, say, the origin $o$. To obtain the average nearest neighbour distribution one would just need to average  $G^u(r)$  with respect to the mean measure~\eqref{e.mean-measure} over  the entire observation window.

For the Mat{\'e}rn II  model, the support of the nearest neighbour distribution $G(r)$ has a lower bound of the inhibition radius,  which is reflected in Figure~\ref{NeighDistMaternII}. The nearest neighbour distribution of the (Mat{\'e}rn-fitted) $\Psi$ does not have such a bound, demonstrating how it has a soft-core. Perhaps a better match would be possible by using determinantal kernels that gives stronger repulsion, which has been a recent subject of mathematical study~\cite{biscio2016quantifying}.

\begin{figure}[t]
\begin{minipage}[b]{0.48\linewidth}
\centering
\centerline{\includegraphics[width=1.1\linewidth]{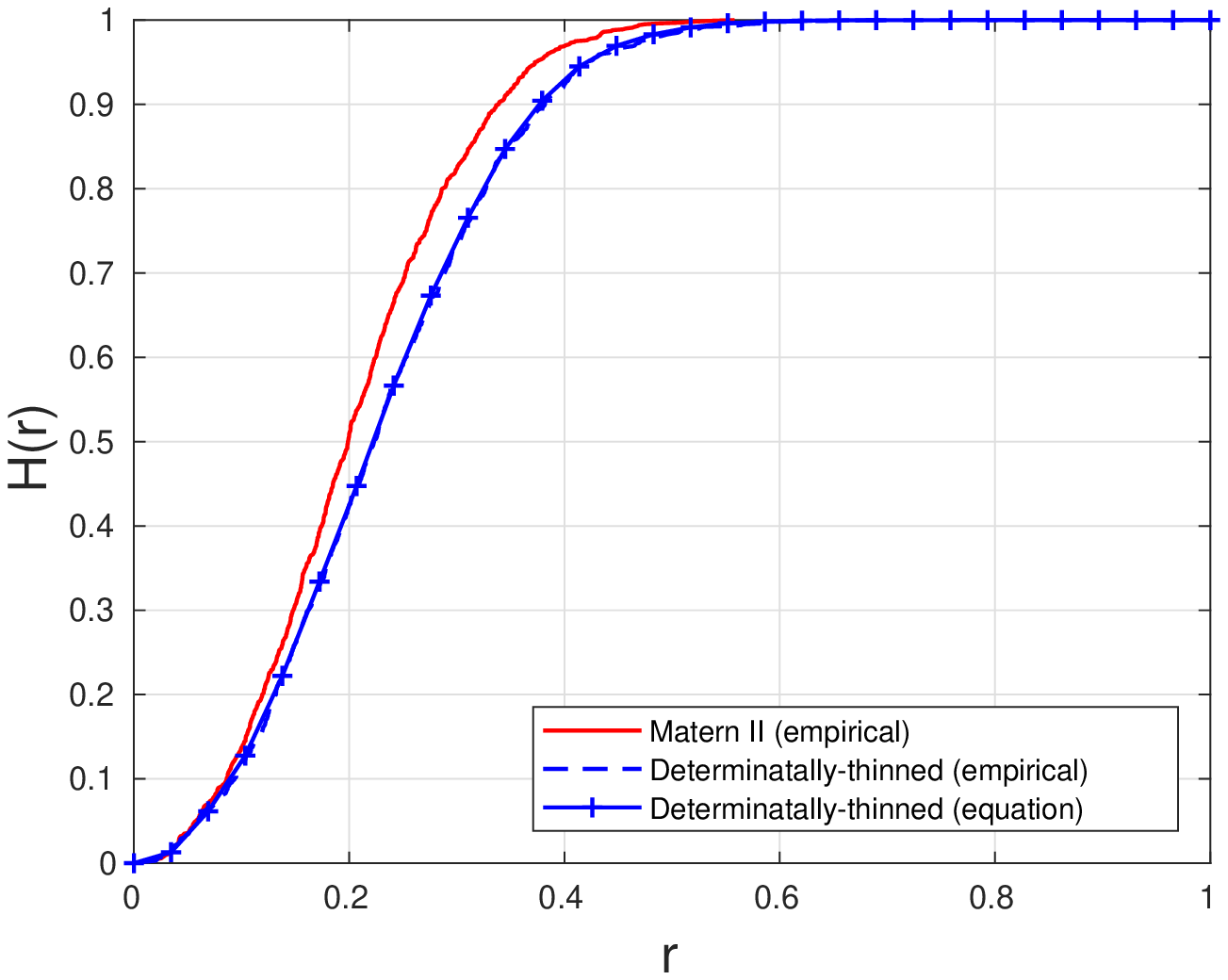}}
\vspace{-2ex}
\caption{\footnotesize 
Empirical spherical contact distribution function of a triangle process and the semi-analytic~\eqref{e.H} with~\eqref{e.void} evaluation of the same function for determinantally-thinned Poisson Poisson process fitted to same the Mat{\'e}rn  II process.}
\label{ContactDistMaternII}
\end{minipage}
\hspace{0.1em}
\begin{minipage}[b]{0.48\linewidth}
\centering
\centerline{\includegraphics[width=1.1\linewidth]{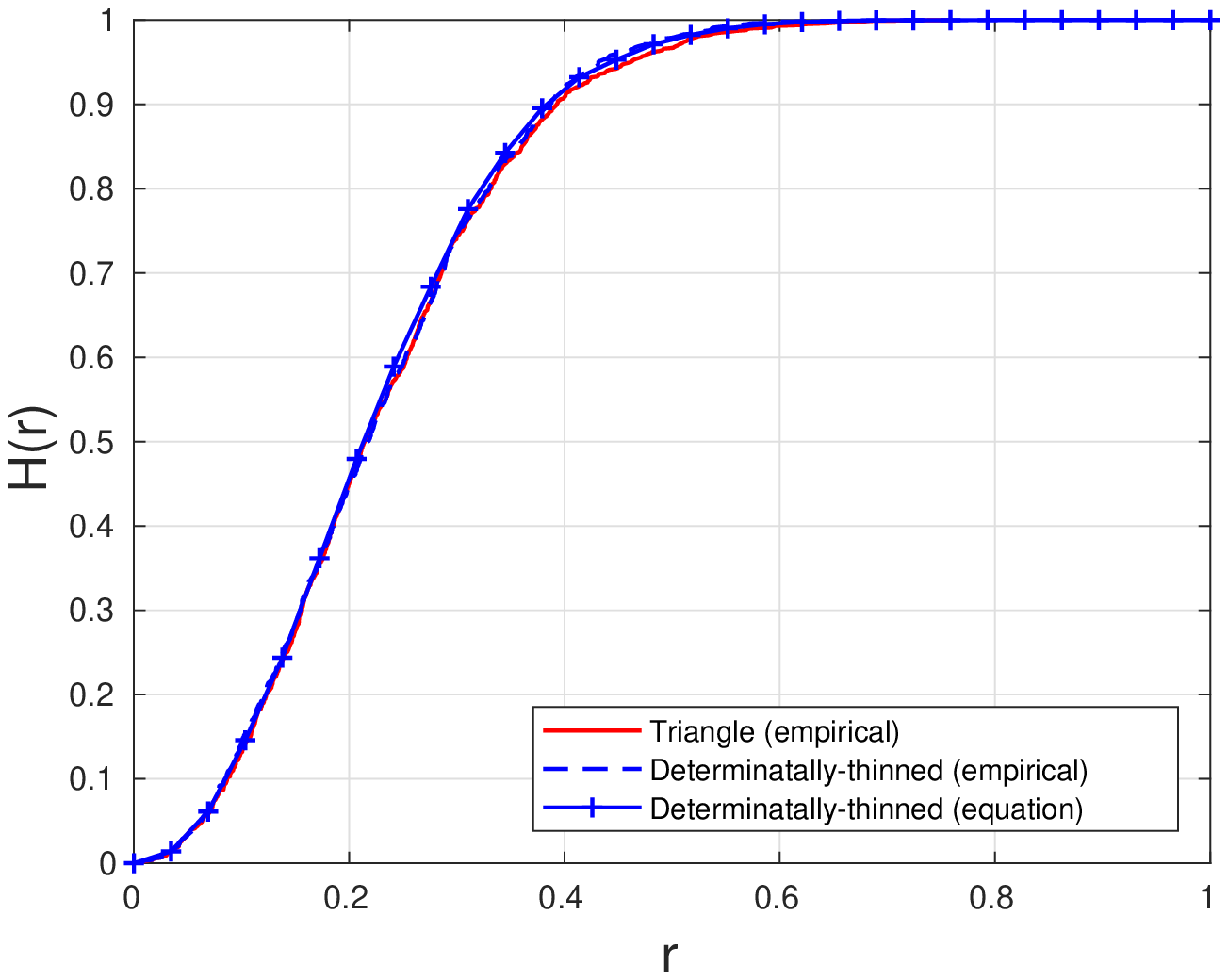}}
\vspace{-2ex}
\caption{\footnotesize Spherical  contact distributions $H(r)$ of a triangle process the fitted determinantally-thinned Poisson Poisson process,
as on Figure~\eqref{ContactDistMaternII}.}
\label{ContactDistTri}
\end{minipage}
\vspace{-3ex}
\end{figure}

\paragraph{Contact distribution} We also studied the (spherical) contact distribution $H_x(r)$, which is the probability distribution of the distance to the nearest point of the point process from an arbitrary location in the region (which we took as the centre of the circular sample window, namely the origin $o$); see ~\cite[Section 8.6.2]{baddeley2015spatial} or~\cite[Section 3.1.7]{chiu2013stochastic}. This is simply the void probability of a disk with radius $r$, so 
\begin{equation}\label{e.H}
H_o(r)=1-\nu_{\Psi}(B_o(r)).
\end{equation}
 (We note that the ratio of the nearest neighbour distribution and this distribution,  known as the $J$ function, is also used as an exploratory test in spatial statistics~\cite[Section 8.6.2]{baddeley2015spatial}). For the distribution $H_o(r)$, edge effects are less of an issue because we study $H_o(r)$ at the centre without conditioning on a point exiting there. Indeed, we see that edge effects have virtually disappeared  for the contact distribution, giving essentially matching curves in Figures~\ref{ContactDistMaternII} and~\ref{ContactDistTri}. 


In summary, the fitted versions of the determinantally-thinned Poisson process $\Psi$ behave statistically like the Mat{\'e}rn II and triangle processes, particularly in  terms of the nearest neighbour and contact distributions. Furthermore, our determinant-based expressions, which only require the Poisson point process to be generated, avoided the statistical bias from edge effects. 

\section{Wireless networks applications}
\label{s.Applications}
\subsection{Models for network layouts}
Many real-world cellular phone network layouts do not resemble realizations of Poisson point processes. When such network layouts exhibit repulsion among the base stations, researchers~\cite{miyoshi2014cellular,torrisi2014large} have proposed using determinantal point processes on the plane $\R^2$ to better model such repulsion.  Though some layouts have been fitted to such point processes~\cite{li2015statistical}, a problem is finding appropriate kernel functions. 
Using models based on determinantal thinning of Poisson  point process circumvents this problem  by allowing one   
to construct the kernels via a very flexible  $L$-ensemble formalism in a way particularly amenable to statistical fitting. One would need to develop a statistical method for fitting complete point patterns in an observation window, and not just their subsets, which would need to address (statistical biasing) issues such as edge effects; see~\cite[5.6.2]{baddeley2015spatial}. 

\subsection{On-off (sleep) schemes}
Instead of modelling network patterns, we now look for appropriate models of subsets of various given network patterns. More specifically, we consider power schemes that power down sections of the networks. A simple model is when each transmitter is independently switched off (or put into sleep mode) with some fixed probability $p$, resulting in a random uncoordinated power scheme. If the original network formed a Poisson point process, then the resulting network of active transmitters forms another Poisson point process. Researchers have used this mathematical model for a power scheme, sometimes called a  blinking process,  to study latency~\cite{dousse2004latency} and routing success in ad hoc networks~\cite{keeler2011model,keeler2012random}, while more recently it has been used to study so-called green cellular networks~\cite{altman2011tradeoffs}.  Although the tractability of such a simple power scheme is appealing, it can result in active transmitters that are clustered relatively close together, giving  unrealistic and inefficient configurations. We believe our determinantal thinning permits for more  realistic models of power schemes. Moreover, the presented expressions for the Palm distribution and the Laplace transform will hopefully allow one to evaluate the performance of such power schemes in terms of semi-explicit expressions for the coverage probabilities based on signal-to-interference-plus-noise ratio  (see~\cite[Chapter~5]{book2018stochastic}), that is, by randomly simulated (or Monte Carlo) evaluation of some  functionals of the underlying Poisson point process $\Phi$, without ever simulating the actual power schemes. 

\subsection{Pre-optimized transmission schedulers}
The previous example of a sleep scheme is just one way to organize wireless network transmissions. Depending on the quantity of interest, there are different optimization goals and methods resulting in different transmission schedulers. The appeal of determinantally-thinned Poisson processes is that they can be readily fitted to these different schedulers. By using the statistical (supervised) learning approach, the benefit is not just  about the  performance evaluation of the  existing schedulers, but also potentially from the algorithmic nature. Imagine the situation in which finding an optimal subset of transmitters requires solving a  computationally heavy problem, so it is not feasible to do online implementations. Instead,  one can solve the optimization problem offline for a sufficiently rich set of configurations of potential transmitters (and receivers) and then use it as a training set to fit a determinantal thinning approximation to the original optimization problem. Such suitable fitted determinantal thinnings could be implemented instead of the original complex scheduler.

\section{Conclusions}
Motivated to present tractable models  for wireless networks exhibiting repulsion, we used determinantal point processes on finite state spaces to define a determinantal thinning operation 
on point processes. The resulting determinantally-thinned point processes possess many appealing properties including accurately fundamental functionals, such as the Laplace functional, void probabilities, as well as Palm (conditional) probabilities, by simulating the underlying (non-thinned) point process, without simulating the actual (thinned) point process. In contrast to Gibbs models, which require weighting the entire realization,  determinantal thinning does not involve intractable normalizing constants, and also, it is  particularly amenable to  statistical  fitting of the parameters. We illustrated this by presenting two examples where the determinantal thinning model is fitted to two different thinning mechanisms that create repulsion. We see them as prototypes for determinantal schedulers  approximating more sophisticated wireless transmission scheduling schemes  with (geometry-based) medium access control, offering new avenues for future research.
 
In this paper we have considered only determinantal thinning of a Poisson point process in a finite window, but completely arbitrary, simple, finite underlying point processes are also possible, including  non-stationary ones on $\R^d$ having finite total number of points.
On a more theoretical note, one can consider the problem of extending this 
setting  to stationary thinning of point processes on~$\R^d$, which raises the question of constructing discrete thinning kernels in random, countable environment;
see~\cite{shirai2003random2} for the theory of determinantal processes on deterministic countable spaces.
Note that the known  kernels used in the continuous setting, such as that of the Ginibre point process,  do not necessarily have required properties when 
defining the discrete operators with respect to, say, Poisson realizations.
Also, the $L$-ensemble approach does not apply directly to the infinite state spaces. A natural extension consists of considering an $L$ with finite dependence range. More precisely, set $[L(\phi)]_{x,y}=0$ for all $\phi$ and  $x,y\in\phi$ such that  $|x-y|>M$ for some constant $M$, meaning that the Gilbert graph with the edge length threshold $M$ does not percolate with probability one on the underlying process (to be thinned). The existence of stationary, determinantal thinning mechanisms with infinite dependence range is left for future fundamental research.

\section*{Acknowledgment}
\addcontentsline{toc}{section}{Acknowledgment
The research for this paper was financially supported 
via the Research Collaboration Agreement
No. HF2016090005 between Huawei Technologies France and Inria
on {\em Mathematical Modeling of 5G Ultra Dense Wireless Networks}.
}
\addcontentsline{toc}{section}{References}

\begin{thebibliography}{10}

\bibitem{altman2011tradeoffs}
E.~Altman, M.~K. Hanawal, R.~ElAzouzi, and S.~Shamai.
\newblock Tradeoffs in green cellular networks.
\newblock {\em ACM SIGMETRICS Performance Evaluation Review}, 39(3):67--71,
  2011.

\bibitem{baddeley2015spatial}
A.~Baddeley, E.~Rubak, and R.~Turner.
\newblock {\em Spatial point patterns: methodology and applications with R}.
\newblock Chapman and Hall/CRC, 2015.

\bibitem{biscio2016quantifying}
C.~A.~N. Biscio and F.~Lavancier.
\newblock Quantifying repulsiveness of determinantal point processes.
\newblock {\em Bernoulli}, 22(4):2001--2028, 2016.

\bibitem{book2018stochastic}
B.~B{\l}aszczyszyn, M.~Haenggi, P.~Keeler, and S.~Mukherjee.
\newblock {\em Stochastic geometry analysis of cellular networks}.
\newblock Cambridge University Press, 2018.

\bibitem{borodin2005eynard}
A.~Borodin and E.~M. Rains.
\newblock Eynard--{M}ehta theorem, {S}chur process, and their {P}faffian
  analogs.
\newblock {\em Journal of statistical physics}, 121(3-4):291--317, 2005.

\bibitem{chiu2013stochastic}
S.~N. Chiu, D.~Stoyan, W.~S. Kendall, and J.~Mecke.
\newblock {\em Stochastic geometry and its applications}.
\newblock John Wiley \& Sons, 2013.

\bibitem{dereudre2017introduction}
D.~Dereudre.
\newblock Introduction to the theory of {G}ibbs point processes.
\newblock {\em arXiv preprint arXiv:1701.08105}, 2017.

\bibitem{dousse2004latency}
O.~Dousse, P.~Mannersalo, and P.~Thiran.
\newblock Latency of wireless sensor networks with uncoordinated power saving
  mechanisms.
\newblock In {\em Proceedings of the 5th ACM international symposium on Mobile
  ad hoc networking and computing}, pages 109--120. ACM, 2004.

\bibitem{gentle2017matrix}
J.~E. Gentle.
\newblock {\em Matrix Algebra: Theory, Computations and Applications in
  Statistics}.
\newblock Springer, 2017.

\bibitem{gomez2015case}
J.-S. Gomez, A.~Vasseur, A.~Vergne, P.~Martins, L.~Decreusefond, and W.~Chen.
\newblock A case study on regularity in cellular network deployment.
\newblock {\em IEEE wireless communications letters}, 4(4):421--424, 2015.

\bibitem{hough2006determinantal}
J.~B. Hough, M.~Krishnapur, Y.~Peres, and B.~Vir{\'a}g.
\newblock Determinantal processes and independence.
\newblock {\em Probability surveys}, 3:206--229, 2006.

\bibitem{keeler2018detpoissoncode}
H.~P. Keeler.
\newblock Simulating and fitting determinantally-thinned {P}oisson point
  processes, 2018.

\bibitem{keeler2011model}
H.~P. Keeler and P.~G. Taylor.
\newblock A model framework for greedy routing in a sensor network with a
  stochastic power scheme.
\newblock {\em ACM Transactions on Sensor Networks (TOSN)}, 7(4):34, 2011.

\bibitem{keeler2012random}
H.~P. Keeler and P.~G. Taylor.
\newblock Random transmission radii in greedy routing models for ad hoc sensor
  networks.
\newblock {\em SIAM Journal on Applied Mathematics}, 72(2):535--557, 2012.

\bibitem{kulesza2010structured}
A.~Kulesza and B.~Taskar.
\newblock Structured determinantal point processes.
\newblock In {\em Advances in neural information processing systems}, pages
  1171--1179, 2010.

\bibitem{kulesza2012determinantal}
A.~Kulesza and B.~Taskar.
\newblock Determinantal point processes for machine learning.
\newblock {\em Foundations and Trends in Machine Learning}, 5(2--3):123--286,
  2012.

\bibitem{kulesza2012arxiv}
A.~Kulesza and B.~Taskar.
\newblock Determinantal point processes for machine learning.
\newblock {\em arXiv preprint arXiv:1207.6083}, 2012.

\bibitem{lavancier2014detextended}
F.~Lavancier, J.~M{\o}ller, and E.~Rubak.
\newblock Determinantal point process models and statistical inference:
  Extended version.
\newblock {\em arXiv preprint arXiv:1205.4818}, 2014.

\bibitem{lavancier2015determinantal}
F.~Lavancier, J.~M{\o}ller, and E.~Rubak.
\newblock Determinantal point process models and statistical inference.
\newblock {\em Journal of the Royal Statistical Society: Series B (Statistical
  Methodology)}, 77(4):853--877, 2015.

\bibitem{li2014fitting}
Y.~Li, F.~Baccelli, H.~S. Dhillon, and J.~G. Andrews.
\newblock Fitting determinantal point processes to macro base station
  deployments.
\newblock In {\em Global Communications Conference (GLOBECOM), 2014 IEEE},
  pages 3641--3646. IEEE, 2014.

\bibitem{li2015statistical}
Y.~Li, F.~Baccelli, H.~S. Dhillon, and J.~G. Andrews.
\newblock Statistical modeling and probabilistic analysis of cellular networks
  with determinantal point processes.
\newblock {\em IEEE Transactions on Communications}, 63(9):3405--3422, 2015.

\bibitem{macchi1975coincidence}
O.~Macchi.
\newblock The coincidence approach to stochastic point processes.
\newblock {\em Advances in Applied Probability}, 7(1):83--122, 1975.

\bibitem{miyoshi2014cellular}
N.~Miyoshi and T.~Shirai.
\newblock Cellular networks with $\alpha$-{G}inibre configurated base stations.
\newblock In {\em The Impact of Applications on Mathematics}, pages 211--226.
  Springer, 2014.

\bibitem{nakata2014spatial}
I.~Nakata and N.~Miyoshi.
\newblock Spatial stochastic models for analysis of heterogeneous cellular
  networks with repulsively deployed base stations.
\newblock {\em Performance Evaluation}, 78:7--17, 2014.

\bibitem{shirai2003random1}
T.~Shirai and Y.~Takahashi.
\newblock Random point fields associated with certain {Fredholm} determinants
  {I}: fermion, {P}oisson and boson point processes.
\newblock {\em Journal of Functional Analysis}, 205(2):414--463, 2003.

\bibitem{shirai2003random2}
T.~Shirai and Y.~Takahashi.
\newblock Random point fields associated with certain {Fredholm} determinants
  {II}: fermion shifts and their ergodic and {G}ibbs properties.
\newblock {\em The Annals of Probability}, 31(3):1533--1564, 2003.

\bibitem{torrisi2014large}
G.~L. Torrisi and E.~Leonardi.
\newblock Large deviations of the interference in the {G}inibre network model.
\newblock {\em Stochastic Systems}, 4(1):173--205, 2014.

\bibitem{wachinger2015sampling}
C.~Wachinger and P.~Golland.
\newblock Sampling from determinantal point processes for scalable manifold
  learning.
\newblock In {\em International Conference on Information Processing in Medical
  Imaging}, pages 687--698. Springer, 2015.

\end{thebibliography}

\appendix

\subsection{Laplace functional}
\label{s.Appendix-Laplace}
Our probabilistic proof of~\eqref{e.Laplace}  exploits an observation
allowing one to interpret  the Laplace functional of a general point process $\Psi$
$$
\textbf{L}_{\Psi}(f)=\E\,\left[e^{-\sum_{x\in \Psi} f(x) } \right]
=\E\,\Bigl[\prod_{x\in \Psi} e^{-f(x)} \Bigr]
$$
as the probability that an independent thinning of $\Psi$ with position dependent
retention probabilities equal to $1-e^{-f(x)}$ has no points in the whole space.
Such an independent thinning of $\Psi$ introduces the product of retention probabilities 
of $x_1,\ldots,x_n$ as a factor to the  moments measures of $\Psi$; see also~\cite[Proposition A.2.]{lavancier2014detextended}. 
When  $\Psi$ is interpreted as  a determinantal thinning of a Poisson process with kernel $K$, then by combining this factor with the determinant in~\eqref{e.pin}, 
we can see  this independent thinning of $\Psi$ as \emph{another} determinantal thinning of the same Poisson process with a kernel given by~\eqref{e.Kernel-Laplace}.
Using  the void probability expression~\eqref{e.void} concludes the proof of~\eqref{e.Laplace} with~\eqref{e.Kernel-Laplace}.

\subsection{Palm distribution}
\label{s.Appendix-Palm}
For a discrete determinantal point process $\Psi$ on $\statespace$ with kernel $K$ and  a  given subset $\palmset\subset\statespace$,
the distribution of the reduced Palm version  $\bar\Psi^{\palmset}$ of $\Psi$ given $\palmset\subset\lattice$  (that is, conditioned to  contain $\palmset$,  with $\palmset$  removed from
the conditional realization) 
can be simply expressed directly using the defining property~\eqref{e.dpp}
\begin{align*}
\P(\bar\Psi^\palmset\supset\psi)&=\P\left(\Psi\supset\psi|\Psi\supset\palmset\right)\\
&=\frac{\P\left(\Psi\supset(\psi\cup\palmset)\right)}{\P\left(\Psi\supset\palmset\right)}\\
&=\frac{\det (K_{\psi\cup\palmset})}{\det (K_\palmset)},
\end{align*}
for $\psi\cap\palmset=\emptyset$.  Schur's determinant identity  allows one to 
express the  ratio of the determinants on the right-hand side of the above expression 
using the Schur complement of the block $K_\palmset$ in $K_{\psi\cup\palmset}$
\begin{equation}\label{e.Schur}
\P(\bar\Psi^\palmset\supset\psi)=\text{Schur}(K_\palmset,K_{\palmset\cup\psi});
\end{equation}
see, for example,~\cite[Section~3.4]{gentle2017matrix}.

Borodin and Rains~\cite[Proposition 1.2]{borodin2005eynard} derived  the following  characterization of $\bar\Psi^\palmset$ in terms of the  $L$-ensemble characterizing $\Psi$ as in~\eqref{e.dpp-L}: $\bar\Psi^\palmset$ admits $L$-ensemble $L^\palmset$ given by
\begin{align}\label{e.palm-L}
L^\palmset:=\left(\left[(I_{{\palmset}'}+L)^{-1}\right]_{{\palmset}'}\right)^{-1}-I\,, 
\end{align}
provided $\palmset'=\statespace\setminus {\palmset}\not=\emptyset$,
where   $I_{{\palmset}'}$ is the square matrix of the dimension of $\statespace$  which has ones on the diagonal corresponding to all the points in ${\palmset'}$ and zeroes elsewhere. In this case, using~\eqref{e.K-L}, one can derive the following  equivalent form of the kernel $K^\palmset$ of $\bar\Psi^\palmset$
\begin{equation}\label{e.palm-K}
K^\palmset= I - 	\left[(I_{{\palmset}'}+L)^{-1}\right]_{{\palmset}'}.
\end{equation}

In the case of  $\Psi$ being the determinantal thinning of  Poisson process~$\Phi$, 
denoting by  $\Psi^{(n)}$ and $\Phi^{(n)}$ the factorial powers of the respective point processes
and  $\palmset=(x_1,\dots, x_n)$, $d\palmset=d(x_1,\dots, x_n)$
we have for any, say non-negative function $f(\palmset,\phi)$, 
\begin{align}
&\E\Bigl[\int_{\mathcal{R}^n}f(\palmset,\Psi\setminus\palmset)\,\Psi^{(n)}(d\palmset)\Bigr]\nonumber\\
&=\E\Bigl[\int_{\mathcal{R}^n}f(\palmset,\Psi\setminus \palmset)\Ind(\palmset\subset\Psi)\,\Phi^{(n)}(d\palmset)\Bigr]\nonumber\\
&=\E\Bigl[\int_{\mathcal{R}^n}\E[f(\palmset,\Psi\setminus\palmset)\Ind(\palmset\subset\Psi)|\Phi]\,\Phi^{(n)}(d\palmset)\Bigr]\nonumber\\
&=\int_{\mathcal{R}^n}\E\Bigl[\E[f(\palmset,\Psi\setminus\palmset)\Ind(\palmset\subset\Psi)|\Phi\cup\palmset]\Bigr]\,\Lambda^n(d \palmset)\nonumber\\
&=\int_{\mathcal{R}^n}\E\Bigl[\E[f(\palmset,\bar\Psi^\palmset|\Phi]\P(\palmset\subset\Psi|\Phi\cup\palmset)\Bigr]\frac{1}{\pi(\palmset)}\,M^{(n)}(d\palmset),
\label{e.Campbell}
\end{align}
where the third  equality follows from $n$-th order Campbell's formula and Slivnyak's theorem for Poisson process, and the last one from the definition of $\bar\Psi^{\palmset}$ as the reduced Palm version of $\Psi$ on  each realization $\Phi\cup\palmset$, as well as~\eqref{e.moments} for the moment measure~$M^{(n)}$, thus proving~\eqref{e.Palm-T}.

Observe that the  conditional distribution of $\bar\Psi^\palmset$ in~\eqref{e.Campbell} given~$\Phi=\phi$ is given by~\eqref{e.Schur} 
when considering the discrete setting $\statespace=\phi\cup\palmset$, $\phi\cap\palmset=\emptyset$, with
$K_\palmset=K_{\palmset}(\phi\cup\palmset)$, $K_{\palmset\cup\psi}=K_{\palmset\cup\psi}(\phi\cup\palmset)$
for $\psi\subset\phi$.
Considering $L=L(\phi\cup\palmset)$ and $K^\palmset=K^\palmset(\phi)$, $L^\palmset=L^\palmset(\phi)$, 
the expressions~\eqref{e.palm-L} and~\eqref{e.palm-K} apply as well.

\end{document}